\pdfoutput=1

\documentclass[11pt]{article}
\usepackage[]{emnlp2021}

\usepackage{times}
\usepackage{latexsym}

\usepackage[T1]{fontenc}

\usepackage[utf8]{inputenc}

\usepackage{microtype}
\usepackage{graphicx}
\usepackage{subfigure}
\usepackage{ulem}
\usepackage{caption}
\usepackage{CJKutf8}
\usepackage{multirow}
\usepackage{amsmath}
\usepackage{makecell}
\usepackage{comment}
\usepackage{array}
\usepackage{booktabs}
\title{Positively transitioned sentiment dialogue corpus for developing emotion-affective open-domain chatbots}

\newcommand*{\affaddr}[1]{#1} 

\newcommand*{\email}[1]{\texttt{#1}}

\author{%
Weixuan Wang\footnotemark[1],  Wei Peng\footnotemark[1]  \footnotemark[2], Chong Hsuan Huang, Haoran Wang \\
\affaddr{Artificial Intelligence Application Research Center, Huawei Technologies}\\
\email{\{peng.wei1\}@huawei.com}\\%
}

\date{}

\begin{document}
\begin{CJK}{UTF8}{gbsn}
\maketitle

\renewcommand{\thefootnote}{\fnsymbol{footnote}}
\footnotetext[1]{Co-first authors.}
\footnotetext[2]{Corresponding author.}
\renewcommand{\thefootnote}{\arabic{footnote}}

\begin{abstract}
In this paper, we describe a data enhancement method for developing Emily, an emotion-affective open-domain chatbot. The proposed method is based on explicitly modeling positively transitioned (PT) sentiment data from multi-turn dialogues. We construct a dialogue corpus with PT sentiment data and will release it for public use. By fine-tuning a pretrained dialogue model using the produced PT-enhanced dialogues, we are able to develop an emotion-affective open-domain chatbot exhibiting close-to-human performance in various emotion-affective metrics. We evaluate Emily against a few state-of-the-art (SOTA) open-domain chatbots and show the effectiveness of the proposed approach. The corpus is made publicly available. \footnote{We are in the process of releasing the corpus at https://github.com/Vicky-Wil/Emily}

\end{abstract}
\section{Introduction}
Developing dialogue systems capable of responding to human emotions has emerged as a major research stream to enhance human engagement in conversation. These research works focused on developing conversational agents perceiving and expressing emotions, for example, ``empathetic listeners/chatbots'' \cite{moel}, \cite{caire}, and ``emotional chatting machines'' \cite{emotional}. While the studies mentioned above focused on generating empathetic responses from the system side, another stream of research works relating to emotion elicitation commenced exploring the effects of the agent's responses on users' emotion states \cite{example}, \cite{elicit}, \cite{emoelicit}. A concurrent research work addressing negative human emotions in dialog systems has recently coined the notion of ``Emotional Support Conversation (ESC)'' \cite{huang}, in which a task framework of ESC is defined. The ESC framework is pivoted at tracking, modeling users' emotional states dynamically, and developing emotional support response generation strategies. This paper describes a data-driven approach for developing an emotion-affective open-domain chatbot (aka ``Emily'') built upon dialogue contexts enhanced with positively-transitioned emotion states across turns.

\begin{figure*}[htbp]

\centering
\subfigure[The emotional state transition diagram of IEMOCAP dataset.]{
\includegraphics[width=7cm,height=7cm]{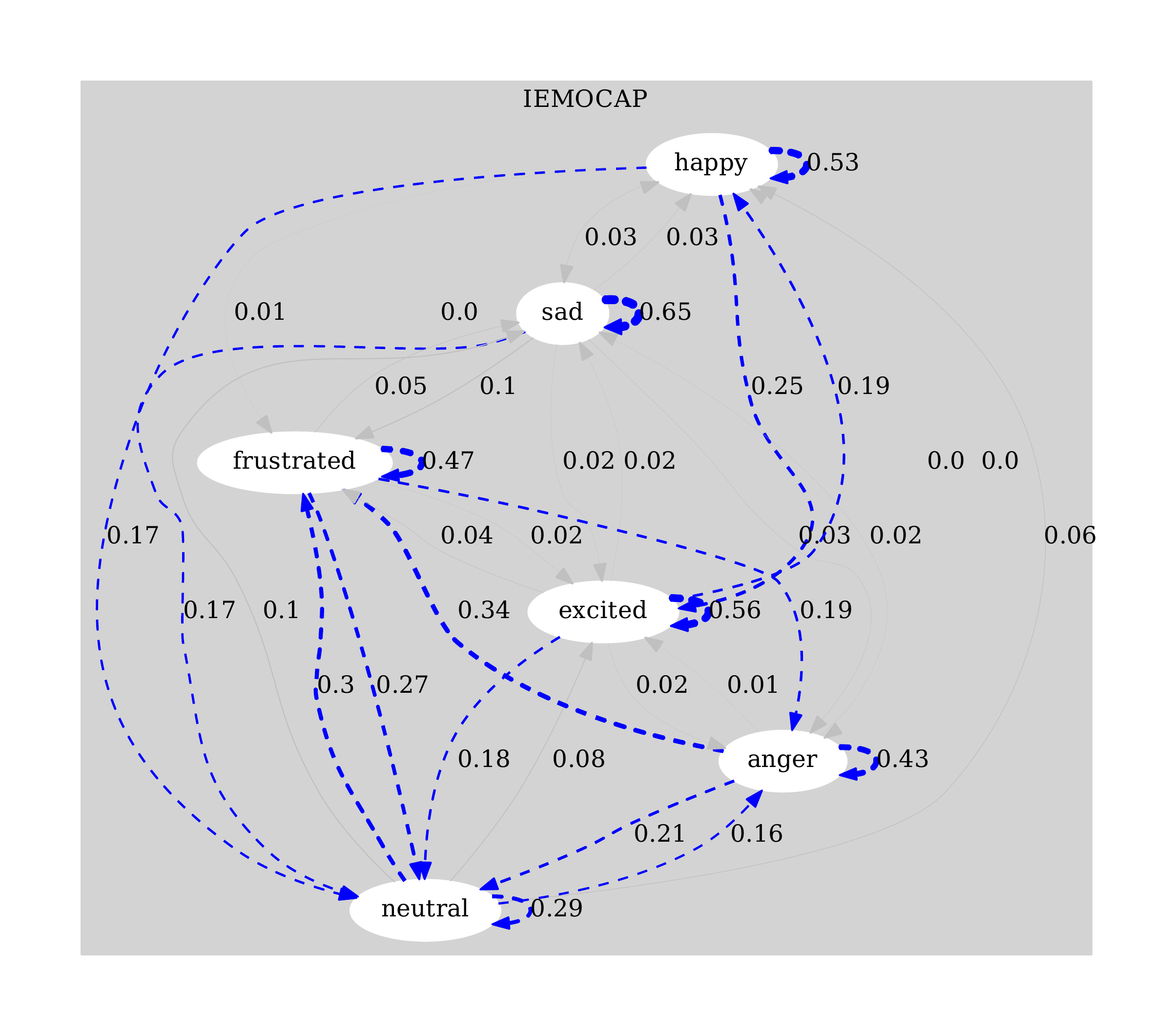}
\label{iemocap-trans}
}
\quad
\subfigure[The emotional state transition diagram of MELD dataset.]{
\includegraphics[width=7cm,height=7cm]{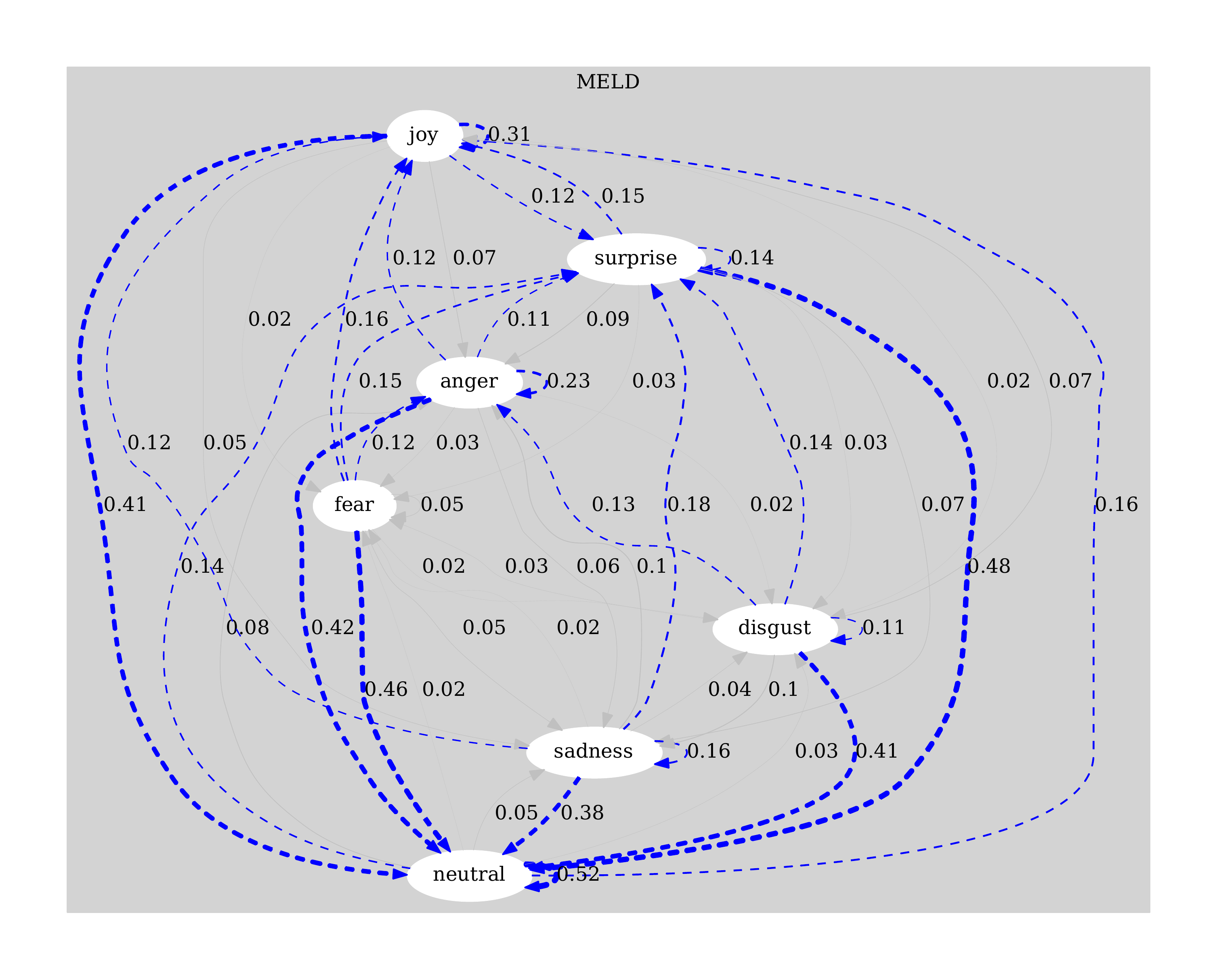}
\label{meld-trans}
}
\caption{The emotional state transition diagrams of IEMOCAP and MELD. The arrows in blue represent state transitions with a probability greater than 0.1, and the arrows in grey denote state transitions with a probability less than or equal to 0.1.}
\label{all}

\end{figure*}

Existing open-domain chatbots are trained using human dialogue datasets endowed with emotions. They are trained end-to-end using all data without tracking or explicitly emphasizing data with specific emotional states transition. We believe that not all dialogue data are equal in relation to developing emotion-affective dialogue systems. We hypothesize that some dialogue data with positive emotional state transitions (i.e., dialogue data with the user's emotional states changed from negative or neutral to positive during conversation) are valuable. As shown in the emotional state transition diagrams of the two popular dialogue datasets IEMOCAP \cite{iemocap} and MELD \cite{meld}, a positive utterance (i.e., ``happy'' in Figure \ref{iemocap-trans}, ``joy'' in Figure \ref{meld-trans}) is most likely to elicit a positive one. The same pattern applies to the negative ones in the opposite camp. It is noted that both positively transitioned and negatively shifted utterances exist, indicating that users' emotional states can be influenced. It is intuitive to extract positively transitioned dialogue data to enhance an open-domain chatbot. To this end, we propose a data enhancement method to develop an emotion-affective chatbot with the aforementioned positively transitioned (PT) sentiment data.

The contributions of this work are as follows:

\begin{itemize}
\item We propose a data enhancement method to explicitly extract positive sentiment state transitions across dialogue turns and, as a result, release a corpus for the public use;   
\item We further develop an emotion-affective open-domain chatbot (Emily) by fine-tuning a DialoGPT model based on the dialogue turns enhanced with positively transitioned (PT) data;
\item We evaluate Emily against a few state-of-the-art (SOTA) open-domain chatbots and show the effects of the proposed approach.
\end{itemize}

\section{Related Work}

Emotion-aware chatbot has become an emerging area of research in recent years. \cite{emotional} first addressed the emotion factor in large-scale dialog generation using an end-to-end framework to generate contents and emotions. \cite{mojitalk} proposed a sophisticated CVAE-based model called ``MOJITALK'', which used emoji to control the emotion and sentiment of the generated responses. \cite{caire} presented an empathetic chatbot ``CAiRE'' which fine-tunes a large-scale pretrained language model with multiple objectives aiming at detecting dialogue emotion and generating empathetic responses. \cite{benchmark} focused on empathetic dialogue generation and released a novel empathetic dialogue dataset as a benchmark. \cite{happy} trained a sentiment predictor with a reinforcement learning framework to encourage more empathetic responses. \cite{moel} introduced a novel dialogue system named ``MoEL'' to perceive the users' feelings and respond accordingly by learning specific listeners for each emotional state.

 
\cite{example} utilized many examples of human appraisal in spoken dialogue to elicit a positive emotional impact throughout the interaction. \cite{elicit} built a chat-oriented dialogue system that can dynamically mimic affective human interaction and generate more natural responses aiming at eliciting a more positive emotional impact. \cite{emoelicit} proposed a variational model EmoElicitor to generate responses that can elicit users' specific emotions with the help of a pretrained language model. \cite{huang} defined a task framework for developing ESC to reduce users' emotional distress via a three-stage procedure (exploration, comforting, and action) and supporting strategies. In this paper, we explore a data-driven approach utilizing positively transitioned sentiment dialogue data to develop an emotion-affective open-domain chatbot.

\begin{figure*}[h]

    \centering
    \includegraphics[width=1.0\textwidth,height=0.25\textheight]{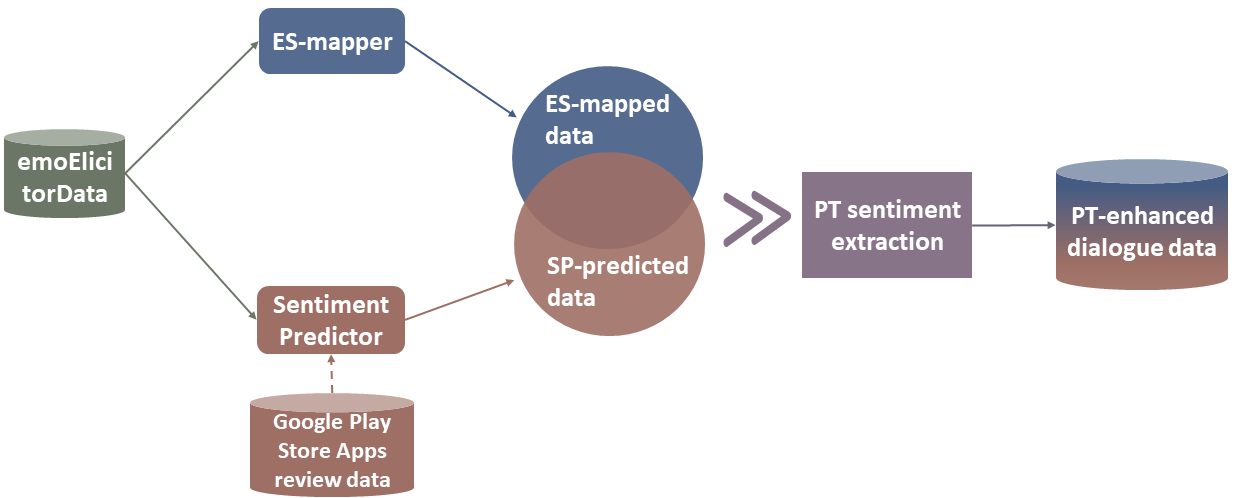}
    \caption{The general flow chart of the data enhancement method producing positively transitioned (PT) sentiment dialogues.}
    \label{flow}

\end{figure*}

\section{Methodology}



The data enhancement process is depicted in Figure~\ref{flow}, in which positively transitioned (PT) dialogue data is extracted. This batch of data is further used to develop the proposed chatbot (Emily) based on the DialoGPT architecture \cite{dialogpt}.

\subsection{Positively Transitioned Sentiment Data Enhancement}
\subsubsection{Training Sentiment Predictor (SP)}
\label{datapre}
The original data (aka EmoElicitorData) used as an input is released from EmoElicitor \cite{emoelicit}, which consists of dialogue turns from Twitter containing emojis. The first task is to produce a sentiment label (positive, neutral, and negative) for each utterance since there is no off-the-shelf multi-turn dialog dataset labeled with sentiments available. We train a sentiment classifier (shown as \textbf{Sentiment Predictor} as Figure~\ref{flow}) by fine-tuning the pretrained language model BERT-base \cite{bert} using the Google Play Store Apps review data available from Kaggle \footnote{https://www.kaggle.com/lava18/google-play-store-apps}. We do not select popular emotion eliciting datasets (IEMOCAP and MELD) because they have skewed sentiment distributions. 

\begin{figure*}[h]

\centering
\includegraphics[width=9cm]{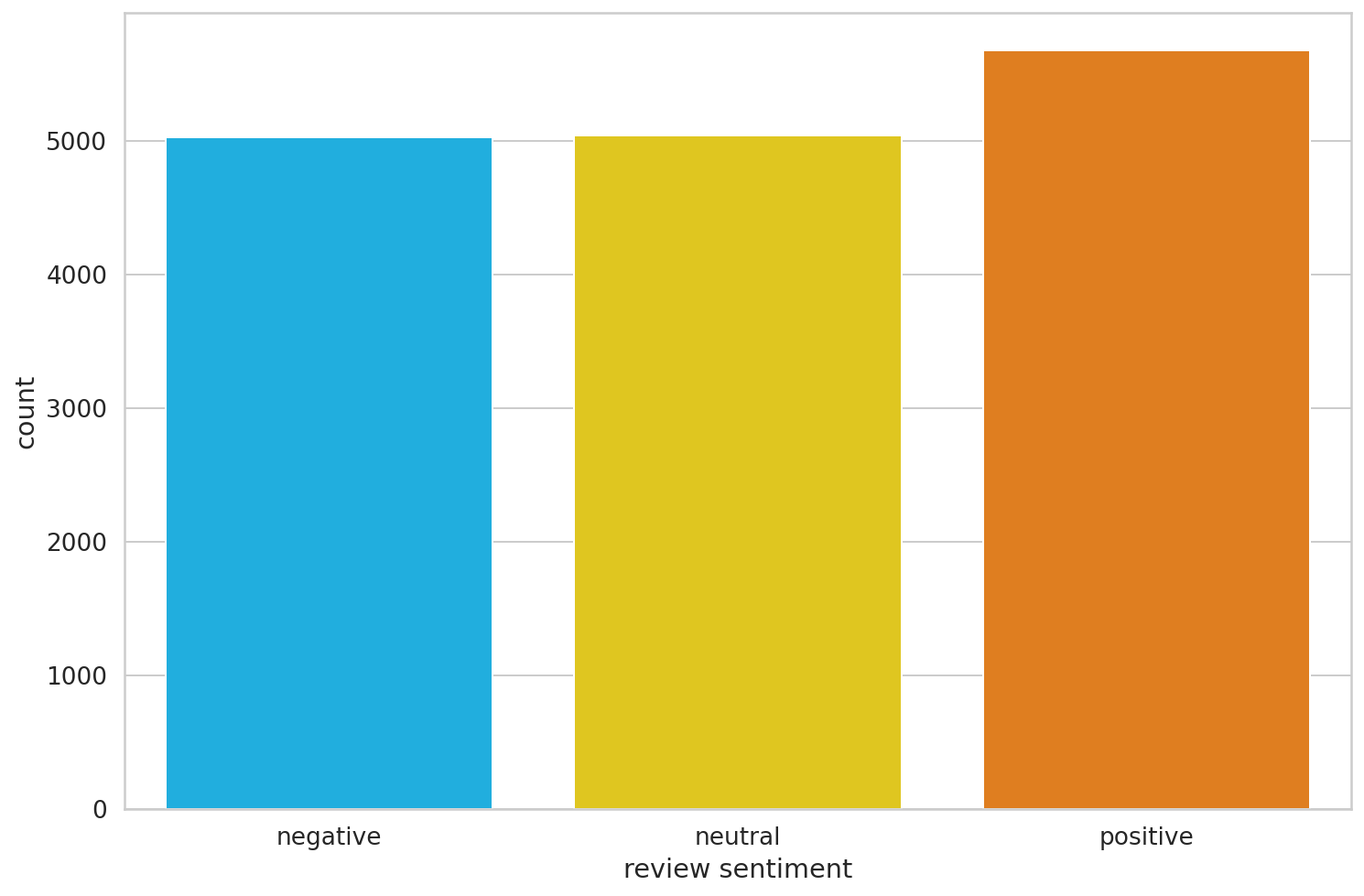}
\caption{\label{distribution}The distribution of sentiment labels for Google Play Store Apps review data.}

\end{figure*}

The review score ranges from 1 to 5, in which we label data with a score <= 2 as negative samples. The data having a score >=4 are categorized as positive samples, with the rest labeled as neutral. The distribution of the review data is shown in Figure~\ref{distribution}. The review data is split into the train, valid, and test sets with a size of 14,171, 787, and 788, respectively. After ten training epochs, the predictor has converged with a test accuracy of $89.59\%$. Other evaluation metrics are shown in Table~\ref{precision}.

\begin{table}[!h]
\centering
\setlength\tabcolsep{3pt}
\begin{tabular}{c|cccc}
\hline
\textbf{Sentiment}  & \textbf{precision} & \textbf{recall} & \textbf{f1-score} & \textbf{support} \\ \hline
\textbf{negative} & 0.91      & 0.87   & 0.89     & 245     \\
\textbf{neutral}  & 0.83      & 0.87   & 0.85     & 254     \\
\textbf{positive} & 0.93      & 0.93   & 0.93     & 289    \\ \hline
\end{tabular}
\caption{\label{precision} The evaluation results of Sentiment Predictor on Google Play Store Apps review data.}
\end{table}

\subsubsection{Developing an Emoji-sentiment Mapper (ES-mapper)}

As the source data (EmoElicitorData) contains multiple emojis, they can be used to reduce the chance of potential mistakenly labeled utterances. We develop an Emoji-sentiment mapper (shown as \textbf{ES-mapper} in Figure~\ref{flow}) to assign sentiment labels based on the associated emojis. We map 1,008 standard emojis to the three sentiment states with some examples shown in Table~\ref{emoji}. When multiple types of emojis appear in a dialogue utterance, the most frequent emoji is chosen to represent the sentiment state of the sentence. 

\begin{table*}[!h]
\centering
\setlength\tabcolsep{5pt}
\begin{tabular}{c|p{280pt}}
\hline
\specialrule{0em}{1pt}{1pt}
\textbf{Sentiment}  &  \textbf{Emoji} \\ \hline
\specialrule{0em}{1pt}{1pt}
\textbf{positive} & \begin{minipage}[b]{0.12\columnwidth} \raisebox{-.\height}{\includegraphics[width=0.4\linewidth]{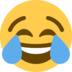}} \end{minipage}  \begin{minipage}[b]{0.12\columnwidth} \raisebox{-.\height}{\includegraphics[width=0.4\linewidth]{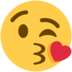}} \end{minipage} 
\begin{minipage}[b]{0.12\columnwidth} \raisebox{-.\height}{\includegraphics[width=0.4\linewidth]{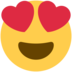}} \end{minipage}
\begin{minipage}[b]{0.12\columnwidth} \raisebox{-.\height}{\includegraphics[width=0.4\linewidth]{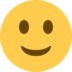}} \end{minipage}
\begin{minipage}[b]{0.12\columnwidth} \raisebox{-.\height}{\includegraphics[width=0.4\linewidth]{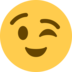}} \end{minipage}
\begin{minipage}[b]{0.12\columnwidth} \raisebox{-.\height}{\includegraphics[width=0.4\linewidth]{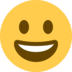}} \end{minipage}
\begin{minipage}[b]{0.12\columnwidth} \raisebox{-.\height}{\includegraphics[width=0.4\linewidth]{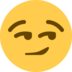}} \end{minipage}
\begin{minipage}[b]{0.12\columnwidth} \raisebox{-.\height}{\includegraphics[width=0.4\linewidth]{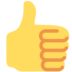}} \end{minipage}
\begin{minipage}[b]{0.12\columnwidth} \raisebox{-.\height}{\includegraphics[width=0.4\linewidth]{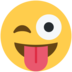}} \end{minipage}
\begin{minipage}[b]{0.12\columnwidth} \raisebox{-.\height}{\includegraphics[width=0.4\linewidth]{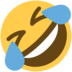}} \end{minipage}
\begin{minipage}[b]{0.12\columnwidth} \raisebox{-.\height}{\includegraphics[width=0.5\linewidth]{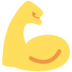}} \end{minipage} \textbf{......} \\ \hline
\specialrule{0em}{1pt}{1pt}
\textbf{neutral}  & \begin{minipage}[b]{0.12\columnwidth} \raisebox{-.\height}{\includegraphics[width=0.4\linewidth]{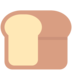}} \end{minipage} 
\begin{minipage}[b]{0.12\columnwidth} \raisebox{-.\height}{\includegraphics[width=0.4\linewidth]{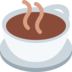}} \end{minipage} 
\begin{minipage}[b]{0.12\columnwidth} \raisebox{-.\height}{\includegraphics[width=0.4\linewidth]{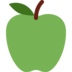}} \end{minipage}
\begin{minipage}[b]{0.12\columnwidth} \raisebox{-.\height}{\includegraphics[width=0.4\linewidth]{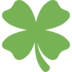}} \end{minipage}
\begin{minipage}[b]{0.12\columnwidth} \raisebox{-.\height}{\includegraphics[width=0.4\linewidth]{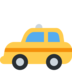}} \end{minipage}
\begin{minipage}[b]{0.12\columnwidth} \raisebox{-.\height}{\includegraphics[width=0.4\linewidth]{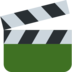}} \end{minipage}
\begin{minipage}[b]{0.12\columnwidth} \raisebox{-.\height}{\includegraphics[width=0.4\linewidth]{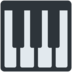}} \end{minipage}
\begin{minipage}[b]{0.12\columnwidth} \raisebox{-.\height}{\includegraphics[width=0.4\linewidth]{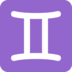}} \end{minipage}
\begin{minipage}[b]{0.12\columnwidth} \raisebox{-.\height}{\includegraphics[width=0.4\linewidth]{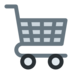}} \end{minipage} 
\begin{minipage}[b]{0.12\columnwidth} \raisebox{-.\height}{\includegraphics[width=0.4\linewidth]{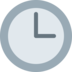}} \end{minipage} 
\begin{minipage}[b]{0.12\columnwidth} \raisebox{-.\height}{\includegraphics[width=0.4\linewidth]{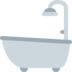}} \end{minipage} \textbf{......} \\ \hline
\specialrule{0em}{1pt}{1pt}
\textbf{negative} & \begin{minipage}[b]{0.12\columnwidth} \raisebox{-.\height}{\includegraphics[width=0.4\linewidth]{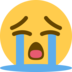}} \end{minipage} 
\begin{minipage}[b]{0.12\columnwidth} \raisebox{-.\height}{\includegraphics[width=0.4\linewidth]{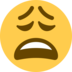}} \end{minipage} 
\begin{minipage}[b]{0.12\columnwidth} \raisebox{-.\height}{\includegraphics[width=0.4\linewidth]{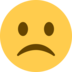}} \end{minipage} 
\begin{minipage}[b]{0.12\columnwidth} \raisebox{-.\height}{\includegraphics[width=0.4\linewidth]{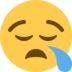}} \end{minipage} 
\begin{minipage}[b]{0.12\columnwidth} \raisebox{-.\height}{\includegraphics[width=0.4\linewidth]{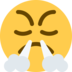}} \end{minipage} 
\begin{minipage}[b]{0.12\columnwidth} \raisebox{-.\height}{\includegraphics[width=0.4\linewidth]{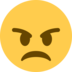}} \end{minipage} 
\begin{minipage}[b]{0.12\columnwidth} \raisebox{-.\height}{\includegraphics[width=0.4\linewidth]{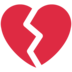}} \end{minipage} 
\begin{minipage}[b]{0.12\columnwidth} \raisebox{-.\height}{\includegraphics[width=0.4\linewidth]{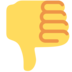}} \end{minipage} 
\begin{minipage}[b]{0.12\columnwidth} \raisebox{-.\height}{\includegraphics[width=0.4\linewidth]{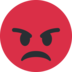}} \end{minipage} 
\begin{minipage}[b]{0.12\columnwidth}
 \raisebox{-.\height}{\includegraphics[width=0.4\linewidth]{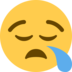}} \end{minipage} 
\begin{minipage}[b]{0.12\columnwidth} \raisebox{-.\height}{\includegraphics[width=0.4\linewidth]{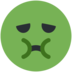}} \end{minipage} 
\textbf{......} \\ \hline

\end{tabular}
\caption{\label{emoji} Examples of emoji-sentiment mapper.}
\end{table*}


\subsubsection{Extracting PT Sentiment Dialogue Data}
\begin{figure*}[h]
\vspace{-0.7cm}
    \centering
    \includegraphics[width=1.0\textwidth,height=0.18\textheight]{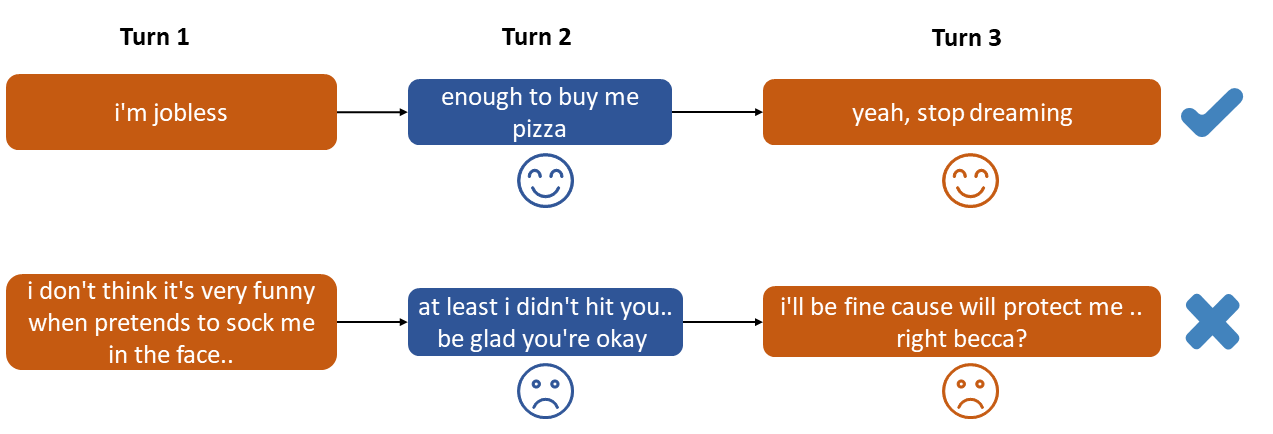}
    \caption{The selection of positively transitioned dialogue turns.}
    \label{select}
\end{figure*}

We adopt a voting mechanism leveraging an agreement of the trained SP and the ES-mapper to generate the final sentiment state of an utterance. The intersection of the sentiment states predicted by SP (shown as \textbf{SP-predicted data} in Figure~\ref{flow}) and the data labeled by ES-mapper (shown as \textbf{ES-mapped data} in Figure~\ref{flow}) is extracted. Based on the voting results of SP and ES-mapper, the original dataset is processed to obtain positive-transition (PT) sentiment data across dialogue turns (shown as \textbf{PT sentiment extraction} in Figure~\ref{flow}). Only dialogue turns with a transition towards a more positive sentiment state are selected (as illustrated in Figure~\ref{select}). 

There are some examples of produced PT dialog (shown in Table~\ref{twitter-exm}). 

\begin{table*}[h]

\centering
\begin{tabular}{c|p{300pt}}
\hline
Human  & Dialogues \\ \hline
A & i'm never gunna be able to sleep tonight scary movies  \\
B & i'll bring you taco bell and protect you     \\
A & perfect girl right here    \\ \hline
A & i have no motivation to do anything today \\
B & we all have days like that. just look at your inspirational pics and your motivation will come :) \\
A & i decided i needed a body to go with the bikinis i just bought lol so i got off the couch \\ \hline
A & yeahh :( i love our family, i'm so sad she's gone :'( \\
B & yes me too i listen the last time. and i want cry because the peoples are wicked with the others. swifties always together \\
A & aww you are so sweet. yes, when i'm sad with my life i know there's here my other family.you guys are so important to me ! \\ \hline
\end{tabular}
\caption{\label{twitter-exm} Examples of PT-enhanced data.}
\end{table*}

\subsubsection{Characteristics of PT sentiment Dialogue Data}
\begin{figure*}[htbp]

\centering
\subfigure[The emotion state transition diagram of PT-enhanced data.]{
\includegraphics[width=5.5cm,height=5.5cm]{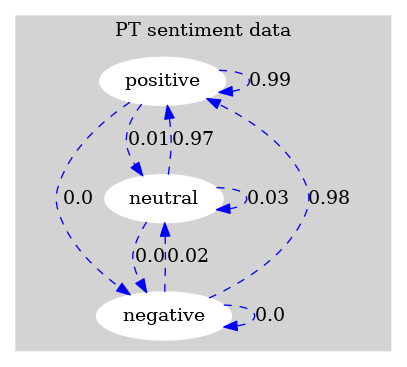}
\label{positive}
}
\quad
\subfigure[The emotion state transition diagram of emoElicitorData.]{
\includegraphics[width=5.5cm,height=5.5cm]{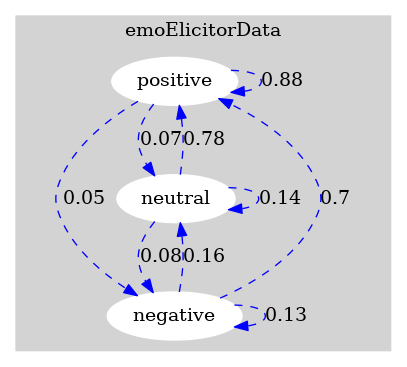}
\label{twitter}
}
\caption{Comparison of the sentiment state transition of PT sentiment data and that of emoElicitorData.}
\label{PT-trans}

\end{figure*}

It can be observed from Figure~\ref{PT-trans} that the PT sentiment dialogue data have distinctive higher probabilities of sentiment state transitions from neural, negative to positive states compared to the original dataset, indicating the effectiveness of the depicted data extraction method.   


Table~\ref{overview} summarizes the main characteristics of the original dataset (emoElicitorData), PT-enhanced data, and Empathetic data. The PT sentiment data contains 67,205 dialogues, which is combined with the Empathetic dataset \cite{benchmark} to fine-tune the model further.

\begin{table*}[]\footnotesize
\centering
\begin{tabular}{l|p{60pt}|p{75pt}|p{55pt}}
\hline
 & \textbf{emoElicitor} & \textbf{PT Sentiment} & \textbf{Empathetic} \\ \hline
\textbf{\# dialogues}                       & 179116                   & 67205                                           & 40250                    \\
\textbf{\# utterances}                    & 651156                   & 302475                                          & 197400                   \\
\textbf{avg \# utterances per dialogue} & 3.6                      & 4.5                                             & 4.9                      \\
\textbf{avg \# words per utterance}   & 11.19                    & 8.56                                            & 9.98                     \\
\textbf{avg \# words per dialogue}      & 40.71                    & 38.5                                            & 48.99 \\ \hline                  
\end{tabular}
\caption{\label{overview}Overview of the dataset of emoElicitorData, PT sentiment dialogue data and the Empathetic data.}

\end{table*}

\subsection{Training an Emotion-affective Chatbot}

The emotion-affective chatbot is developed based on the DialoGPT-large architecture \cite{dialogpt}, initialized with DialoGPT-large model parameters of 762M with 36 layers. The base model is fine-tuned with the selected data mentioned in Section~\ref{datapre} until it converges. Following the same way DialoGPT treats a multi-turn dialogue session, we combine all utterances from a dialogue into a long text sequence $[x_1,...,x_N]$, ending with a special token $<sep>$ \cite{dialogpt}. The dialogue context is denoted as $X=x_1,...,x_k$ and the response to be predicted is $\bar{X}=x_{k+1},...,x_{N}$. The conditional probability $P(\bar{X}|X)$ is calculated as $P(\bar{X}|X)=\prod_{n=k+1}^{N} p(x_n|x_1,...,x_{n-1})$. 


In addition to the model parameters, we learn the word embeddings of the tokens with a size of 1,280. The size of the vocabulary is 50,257. We also use the PT-enhanced data to fine-tune DialoGPT 345M model (reverse, for maximum mutual information \cite{mmi} reranking) \cite{dialogpt}. The purpose of using maximum mutual information (MMI) rescoring is to produce more diverse responses and remove uninformative responses \cite{dialogpt}. After generating a set of hypotheses using Top-K (K=10) sampling, we use the probability produced by MMI to rerank all hypotheses to produce more diverse responses. 

During the training, we set the batch size as 32. The learning rate is 1.5e-4 with a warm-up step number of 2,000. Moreover, we set both the gradient accumulation and maximum gradient normalization to 1.0. The inference time of Emily is around 1.5s per utterance.

\section{Experiments and Evaluation}

In this section, we describe a range of experiments comparing Emily with two SOTA open-domain chatbots, Blender \cite{blender}, and PLATO-2 \cite{plato}, on related metrics.

\subsection{Automatic Evaluation}

\subsubsection{Emotion-affective Response Evaluation Metrics}

We adopt three well-accepted automatic evaluation metrics to compare Emily with Human (reference data), Blender and PLATO-2.

\begin{itemize}
\vspace{-0.3cm}
\item \textbf{Context: }We measure the context correlation between the responses produced by the dialogue model and the queries, similar to a metric used in \cite{metric}.

\item \textbf{Fluency: }Fluency is often measured using a language model. It indicates the negative perplexity of generated responses. We adopt the calculation method proposed by \cite{metric} to produce the fluency score.

\item \textbf{BLEU: }We compute BLEU scores to compare the generated responses against human responses. 

\vspace{-0.3cm}
\end{itemize}


\subsubsection{Automatic Evaluation Results}

\begin{figure*}[htbp]
\centering
\subfigure[Distributions of the context metric of responses recorded for human, Emily, Blender and PLATO-2.]{
\includegraphics[width=9cm]{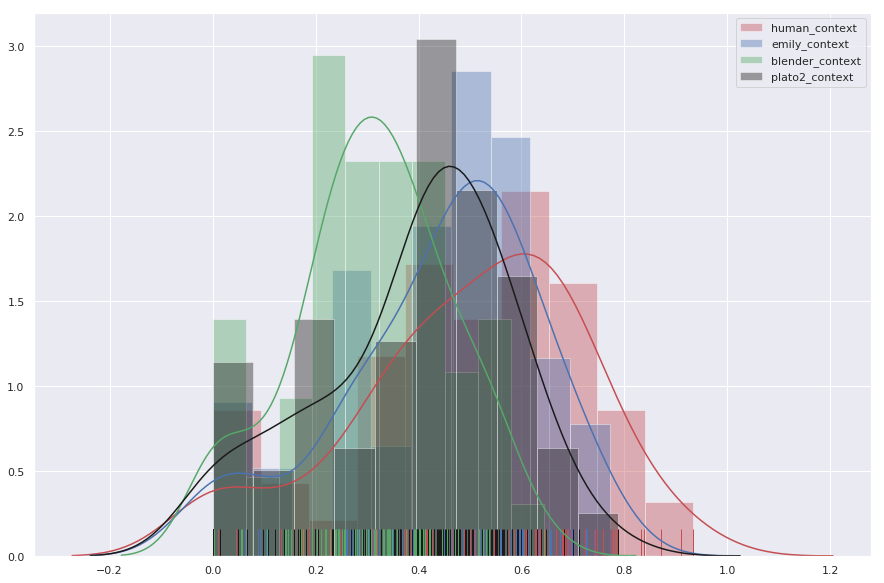}
\label{fig1}
}
\quad
\subfigure[Distributions of the fluency metric of responses recorded for human, Emily, Blender and PLATO-2.]{
\includegraphics[width=9cm]{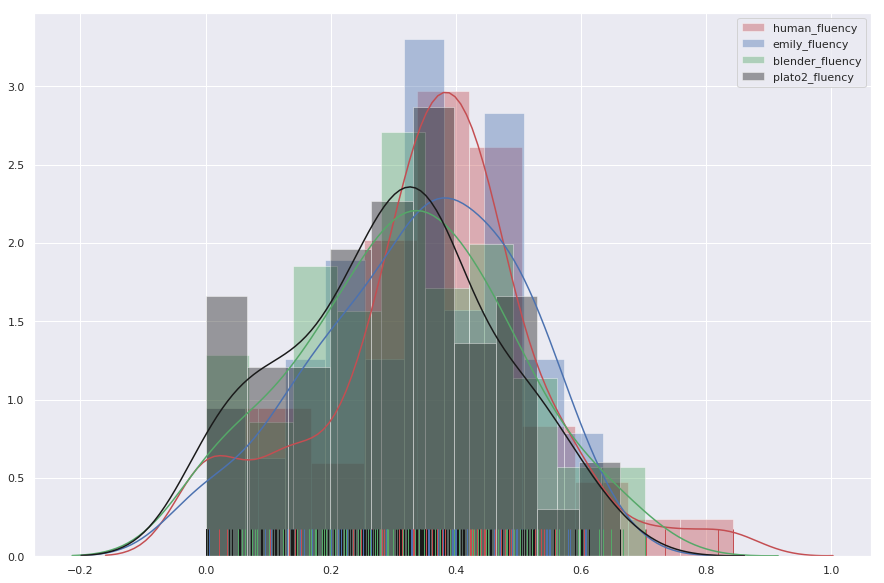}
\label{fig2}
}
\quad
\subfigure[Distributions of the BLEU score of responses recorded for human, Emily, Blender and PLATO-2.]{
\includegraphics[width=9cm]{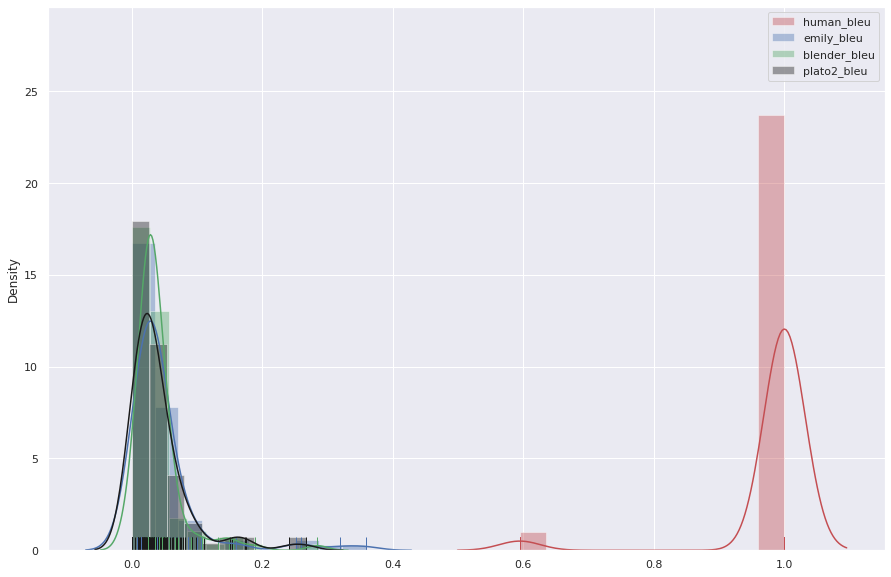}
\label{fig3}
}
\caption{Distribution results on the automatic evaluation metrics.}
\label{all}
\vspace{-0.8cm}
\end{figure*}

We randomly sample 100 dialogues from an empathetic dialogue test dataset \cite{benchmark}. Emily, Blender, and PLATO-2 produce the corresponding response generations with the results shown in Table~\ref{result}. We can see that Emily outperforms Blender and PLATO-2 on every metric (both in average and median) from the results. The distributions of all systems show a consistent result (illustrated in Figure~\ref{all}). It can be observed that Emily (in blue histogram/distribution) has performances closer to human judgments (the red histogram/distribution) than those recorded for Blender and PLATO-2 in both context and fluency metrics. Moreover, Figure~\ref{fig3} shows the distribution of Emily, Blender, and PLATO-2 evaluated on the BLEU score. 

\begin{table*}[h]

\centering
\begin{tabular}{l|p{67pt}p{67pt}p{67pt}p{67pt}}
\hline
Metrics  & Human  & Blender & Plato-2 & \textbf{Emily}  \\ \hline
\textbf{Context} & 0.4977/0.5364 & 0.3098/0.3150  & 0.3918/0.4327  & \textbf{0.4401/0.4881} \\
\textbf{Fluency} & 0.3595/0.3710 & 0.3206/0.3240  & 0.2987/0.3149  & \textbf{0.3398/0.3549} \\
\textbf{BLEU}    & 0.9837/1 & 0.0400/0.0285  & 0.0409/0.0274  & \textbf{0.0464/0.0315} \\ \hline
\end{tabular}
\caption{\label{result} The result of automatic evaluation on context, fluency and BLEU presented in the format of average/median.}

\end{table*}


\subsection{Emotion-affective Exemplars}

Table~\ref{example} demonstrates some responses produced by humans, Emily, Blender, and Plato-2 with regard to a user's utterance, respectively. It can be observed that the dialogue responses Emily generated exhibit a high-level of empathy, relevance, and fluency. However, it is not our intention to claim Emily is superior to any of the involved SOTA open-domain chatbots without a full-scale human evaluation. We will leave this as our future study.

\begin{table*}[h]\small

\setlength\tabcolsep{3pt}
\begin{tabular}{p{105pt}|p{105pt}|p{105pt}|p{105pt}}
\hline
\textbf{Human}   & \textbf{Emily}  & \textbf{Blender} & \textbf{PLATO-2}   \\ \hline
\specialrule{0em}{1pt}{1pt}
\multicolumn{4}{p{340pt}}{\textbf{User}: One of the hardest times in my life was when I had to leave my husband and daughter when I went through a divorce. }  \\ \hline
\specialrule{0em}{1pt}{1pt}
I am so sorry. I can't imagine what you must have went through.    &  Oh no. i am so sorry to hear that happened but it sounds like life's pretty good right now for you and your family?  &  I am so sorry to hear that. I hope you are doing better now. I am divorced too.       & I'm sorry to hear that. why did you divorce?  \\ \hline
\specialrule{0em}{1pt}{1pt}
\multicolumn{4}{p{340pt}}{\textbf{User}: I'm always appalled when I see bullying in action -- whether it's on a playground or political stage -- not much makes me angrier. }  \\ \hline
\specialrule{0em}{1pt}{1pt}
I feel the same way.   & I feel the same way. it is so unfair that someone can be looked down on just because of how you look.. and if they make an effort to change then fine, but don't take things too personally.   & I agree, bullying is a terrible thing. I'm glad you don't let it get to you.     & It's awful! the bullying in my school never stops. \\ \hline
\specialrule{0em}{1pt}{1pt}
\multicolumn{4}{p{340pt}}{\textbf{User}:I feel lonely sometimes cause I live in a house with 5 girls and all of them have boyfriends and sometimes they are all hanging out with their significant others and I feel lonely. } \\ \hline
\specialrule{0em}{1pt}{1pt}
Awww, That sucks man. Hopefully they don't say mean things to you about being lonely.  & I can relate to that. wow, i am sorry you feel lonely!  & I know how you feel. I live with 3 kids and 3 dogs. I run to get away.    &  Maybe you just have to find that special someone who makes you feel wanted and you won't be lonely anymore.   \\ \hline
\specialrule{0em}{1pt}{1pt}
\multicolumn{4}{p{340pt}}{\textbf{User}: I'm always appalled when I see bullying in action -- whether it's on a playground or political stage -- not much makes me angrier.}  \\ \hline
\specialrule{0em}{1pt}{1pt}
I feel the same way.  & I feel the same way. it is so unfair that someone can be looked down on just because of how you look.. and if they make an effort to change then fine, but don't take things too personally.   & I agree, bullying is a terrible thing. I'm glad you don't let it get to you.     & It's awful! the bullying in my school never stops.  \\ \hline
\specialrule{0em}{1pt}{1pt}
\multicolumn{4}{p{340pt}}{\textbf{User}: Ugh! I looked like such a fool at my girlfriend's work party! 
} \\ \hline
\specialrule{0em}{1pt}{1pt}
What happened?     & Oh no! what happened?  & I'm sorry to hear that. I'm sure you didn't look like a fool to her. & Oh no! how could you have looked like a fool? was she having a work party or something? \\\hline
\specialrule{0em}{1pt}{1pt}
\multicolumn{4}{p{340pt}}{\textbf{User}: I can still look back with shame at the way I treated a girl in sixth grade. I wasn't as bad as others, but that doesn't make me feel any better. }    \\ \hline
\specialrule{0em}{1pt}{1pt}
Where you not secure in your self?   &  I know how you feel. i have been burned as a result of my actions at times, and it feels terrible to be that way still.  & I am sorry you had to go through that. I am sure you have learned from your mistakes.  & Aww man, don't beat yourself up. everyone feels that way at one point. what made you treat her so badly? \\ \hline
\specialrule{0em}{1pt}{1pt}
\multicolumn{4}{p{340pt}}{\textbf{User}: I recently had a big argument with my sister. We both got very upset and angry at each other. } \\ \hline
\specialrule{0em}{1pt}{1pt}
Isn't that what sisters are for?  What were you guys upset about?     & I am sorry to hear that. did you make it up with her?     & I'm sorry to hear that. I hope you guys can work it out. What was the argument about?             & What was the argument about? was it about something that happened in a past relationship?     \\ \hline  
\specialrule{0em}{1pt}{1pt}
\multicolumn{4}{p{340pt}}{\textbf{User}: My oldest and my husband have always had a rough relationship. I've got to believe that as she gets older, it will get better! } \\ \hline
\specialrule{0em}{1pt}{1pt}
I'm sure it will! I didn't get along with my dad until i became an adult.    & Absolutely! you must be so proud to see that trust is still unbreakable. i bet it will grow over time for both of y'all.   & I'm sorry to hear that. I hope things get better for you and your family. Do you have any hobbies?            & Yes i believe it will, it sounds like you have been supporting her.     \\ \hline 

\end{tabular}
\caption{\label{example} Examples of dialogue responses generated by human reference, Emily, Blender and PLATO-2 respectively. }

\end{table*}

\section{Conclusion}
In this work, we present a data-driven approach for developing an emotion-affective open-domain chatbot Emily, which is designed to generate emotion-affective responses to a negative sentiment utterance.
The data enhancement method explicitly extracts dialogues with positive sentiment state transitions across turns. By fine-tuning a pretrained dialogue model on positively transitioned (PT) sentiment dialogues, we develop an emotion-affective chatbot achieving close-to-human performances in various emotion-affective metrics. We evaluate Emily against a few SOTA open-domain chatbots and show the effectiveness of the proposed approach. Future work will enhance the emotion affecting the capability of Emily and benchmark the results in more extensive experiment settings.

\bibliography{anthology,custom}
\bibliographystyle{acl_natbib}

\end{CJK}
\end{document}